# The NoN Approach to Autonomic Face Recognition

Willie L. Scott, II
IBM, Austin, Texas

**Abstract**

A method of autonomic face recognition based on the biologically plausible network of networks (NoN) model of information processing is presented. The NoN model is based on locally parallel and globally coordinated transformations in which the neurons or computational units form distributed networks, which themselves link to form larger networks. This models the structures in the cerebral cortex described by Mountcastle and the architecture based on that proposed for information processing by Sutton. In the proposed implementation, face images are processed by a nested family of locally operating networks along with a hierarchically superior network that classifies the information from each of the local networks. The results of the experiments yielded a maximum of 98.5% recognition accuracy and an average of 97.4% recognition accuracy on a benchmark database.

# The NoN Model

We propose an NoN approach which is based on a model of cortical processing for autonomic face recognition. The NoN (network of networks) model was independently developed by Sutton *et al.* [15] and Anderson *et al.* [1]. The model is based on two fundamental ideas in neurobiology: (1) *groups* of interacting nerve cells, or neurons, encode functional information, and (2) processing occurs simultaneously across different *levels* of neural organization. Although like other models of biological neural networks the NoN model makes simplifying assumptions about neural activity, systems using this approach have performed very well.

Nested distributed systems denote the core architecture of the model and they may be seen as having emerged out of an evolutionary process. In the context of describing computational features of the cerebral cortex, this organizing principle was first proposed by Mountcastle [9]. Sutton *et al.* [15] extended this work by providing formal mathematical structure to the proposed theory, whereas Anderson *et al.* [1] investigated NoN in the context of signal processing and how the brain sorts out noisy information. Recently, joining the two approaches [3,16]. Sutton and Anderson applied the NoN model to the analysis and classification of multiple radar signals that were received simultaneously.

The formation of clusters and levels among neurons is based on their interconnections [3]. The NoN model suggests that despite enormous diversity in the connection patterns associated with individual neurons, many neural circuits can be subdivided into essentially similar sub-circuits, where each sub circuit contains many types of neurons. This hierarchy is evident in the cerebral cortex, which is the most complex and elusive of neural circuits [20]. (See Figure 1 adapted from [16]).

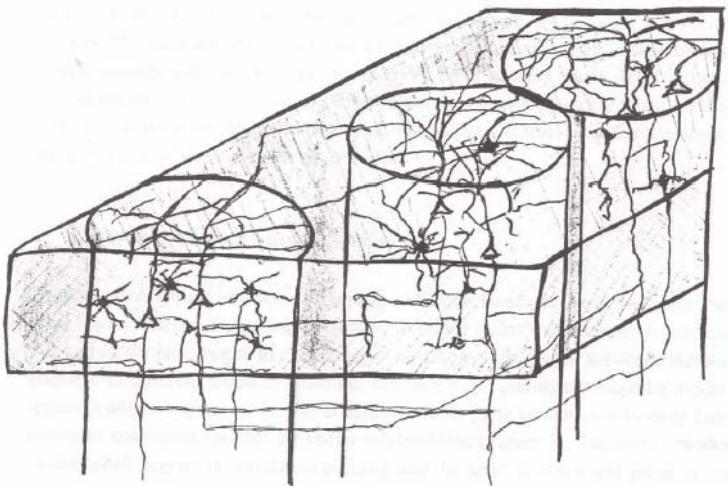

Figure 1. Schematic representation of the cerebral cortex. Three networks of intermediate level organization are displayed.

This nesting arrangement serves to link different and often widely separated regions of the cortex in a precise but distributed manner. Several physiological responses, such as those occurring in the visual cortex in response to optical stimuli, may be associated with each first-level sub-circuit [15]. The result is that nested clusters and their associated memory properties are organized in a complex and distributed manner at each level of the hierarchy (See Figure 2).

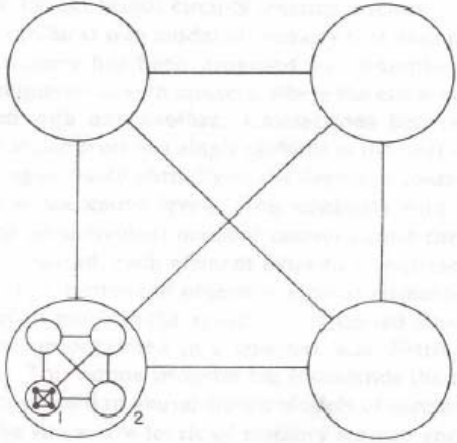

Figure 2. Hierarchy of neural clusters demonstrating levels of organization. Clusters at the first level contain neural elements (small dots) which are fully interconnected with each other, as shown in cluster 1. Certain linkages between first-level clusters form second-level clusters, such as cluster 2.

The NoN method has been used, with excellent results, for image regularization [3], a processing technique which attempts to alleviate the degradations and provide clear and noise-free images, and robotics by Sutton and Anderson, and their associates. In this paper we extend the application of this biologically plausible approach to the problem of face recognition.

# Hierarchical Processing of Face Images

The application of the NoN approach is well suited for the problem of face recognition, due to the analogous manner in which faces are processed in the human vision system (HVS). The human ability to recognize faces efficiently and accurately, as well as other forms of visual stimuli, is facilitated by *spatial vision*. Spatial vision is considered perhaps the most important aspect of human sight, which is defined as the ability to see external objects and discriminate their different shapes. In early spatial vision research, the concept of spatial frequency orientation selectivity was shown to be correct in the HVS. This important concept is captured by the contrast sensitivity function (CSF), a transfer characteristic used to model the HVS as a measure of the organism's response to various spatial frequencies. The CSF displays that the highest sensitivity is in the mid spatial frequency range, with a drop in sensitivity to high spatial frequencies, and a gentler but still pronounced loss in sensitivity at low spatial frequencies as well, when frequency is plotted on a logarithmic scale (See Figure 3).

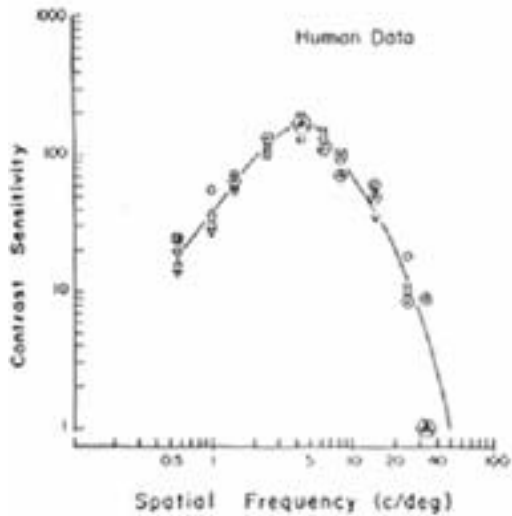

Figure 3. A typical CSF for a normal human observer. The units of spatial frequency are cycles per degree.

The hierarchical processing may be implemented by the DCT, a spatial frequency transform. The DCT separates the image into parts (or spectral sub-bands) of differing importance (with respect to the image's visual quality). For most images, much of the signal energy lies at low frequencies (corresponding to large DCT coefficient magnitudes); these are relocated to the upper-left corner of the DCT. Conversely, the lower-right values of the DCT array represent higher frequencies, and turn out to be smaller in magnitude, especially as $u$ and $v$ approach the subimage width and height, respectively. The DCT is done is small contiguous blocks. In general, hierarchy of blocks of different sizes may be processed to provide an implementation of a mjlti-layered NoN approach.

The input image of a face is typically modeled as a two dimensional array of numbers, i.e., pixel values. It may be written as $X = \{x_i, i \in S\}$ where $S$ is a square lattice. It may be more convenient to express $X$ as a one dimensional column vector of concatenated row of pixels, $X = [x_1, x_2, ..., x_N]^T$ where $N$ is the total number of pixels in the image. For images of size 92x112, which is the size of images used in this study, $N$ is as large as 10,304. Generally speaking, very high dimensional features are usually inefficient and also lack discriminating power. To handle the high dimensionality of the features, we must transform X into a *feature* *vector* $f(X) = [f_1(X), f_2(X), ..., f_M(X)]^T$ where *f₁(x), f₂(x),...,fm(x)* are linear or non-linear functionals. Generally *M* is required to be much smaller than *N* in order to increase the efficiency of the new representation.

The features are representative characteristics extracted from annotated objects to be classified in a recognition system. In the NoN systems, these features are grouped in a 1-D array, forming a feature vector.

In the general NoN case, an *n*-level hierarchy of nested distributed networks is constructed; in the proposed system we take *n* to be 2. The proposed system processes input face images by a nested family of locally operating networks along with a hierarchically superior network that classifies the information from each of the local networks.

First, an input face image is divided into *N x N* blocks to define the local regions of processing. We let *N* be even, i.e., $N = 2^m$, where *m* > 2, due to the computational benefits of doing so in the DCT. In dividing a face image into *N* x *N* blocks, if the image dimensions do not divide by the subimage size, the image is zero-padded to the next multiple that will allow for division. Next, the *N x N* 2-D DCT is used to transform the data into the frequency domain. The transformed version of the face is supplied to Level 1. Level 1 is composed of a nested family of locally operating networks that calculate various functions of spatial frequency in the block, producing a block-level DCT coefficient, which represents a compacted form of the spatial information in the block. At this point, the image is now transformed into a variable length vector and fed to Level 2. Level 2 is a hierarchically superior classifier that is trained with respect to the data set. The output of Level 2 is the classification decision (See Figure 4).

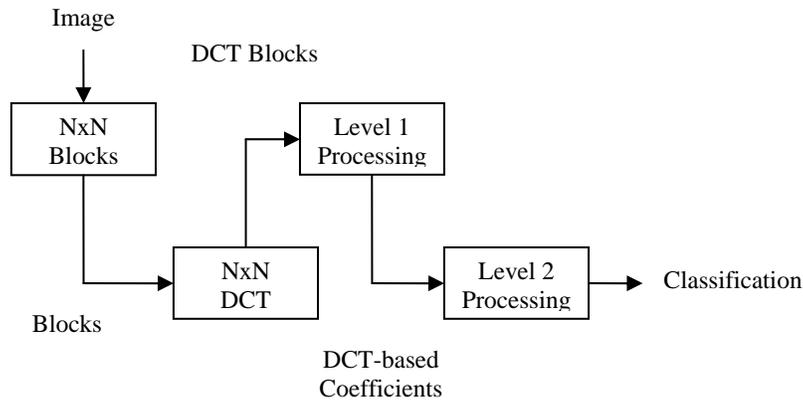

Figure 4. A block diagram view of the face recognition system.

# Nested Family of Locally Operating Networks

Level 1 of the NoN system is a nested family of locally operating networks that calculate various functions of spatial frequency in a block, referred to as block-level DCT coefficients, which are biologically inspired by the CSF. An important property of an algorithm for computing block DCT coefficients is high-level energy compaction. In high-dimensional data, such as face images, the ability to access, store, and transmit information in an efficient manner is important. In a face recognition system, to utilize images effectively, actions must be taken to reduce the number of dimensions required for their representation. In general, a block-level DCT algorithm should further compact the spatial energy of the produced set of transform coefficients into a single representative block-level coefficient.

Perhaps the most essential property of a potential algorithm is the ability to operate on the spatial frequency information in a set of transform coefficients, in order to produce features that can be used for recognition in a manner analogous to the human vision system. As discussed before, the human vision system combines the configuration cues of a face (lower spatial frequencies) with additional cues about the internal features and other details of the face (higher spatial frequencies) in a hierarchical manner, in performing face recognition. As a biologically motivated scheme, an algorithm for computing a block-level coefficient over a set of transform coefficients should also use cues from the entire spatial frequency spectrum in producing a representative coefficient.

# Biologically Motivated Algorithms

In the NoN system, we initially divide the input image into $N \times N$ blocks. We then apply the $N \times N$ 2-D DCT to all subimages (blocks) of a face image to reduce the information redundancy, and use a packed, vector form of the image in classification. To be more precise, let $C \in \Re^{NxN}$ be the 1-D vector representation, in raster scan order (i.e., row-major), of the $N \times N$ 2-D DCT coefficients for a block. We define a function $\phi : \Re^{NxN} \to \Re$ as a function over the vector $C$. Ultimately, we present DCT-based feature vector representations of face images, based on varying the function $\phi$, for use in classification.

In all of the subsequent algorithms for computing block-level DCT coefficient features, the summation step for energy compaction over the DCT block begins at 2 in the 1-D raster scan order. This is because the 2-D DCT places the DC (Direct Current) coefficient in the [0,0] position of the DCT array, and for frequency based calculations, luminance AC (Alternating Current) coefficients provide the most information concerning frequency information in the DCT block. Observing only the AC coefficients of the DCT block is reasonable, in that they convey the most useful information for use in spatial frequency energy compaction.

**Direct Accumulation**

The algorithms in this sub-section accomplish high-level spatial energy compaction, while creating block-level DCT coefficient features that are biased towards the lower frequency spectrum. This biasing is motivated by the human capacity for face recognition of blurred or noisy images.

The first scheme proposed for computing the block-level DCT coefficient, denoted as $\lambda$, is based on taking $\lambda$ as the squared sum of the DCT coefficients $c_i$ in the block. Specifically, we let $C$ be the set of DCT coefficients of a $N \times N$ block arranged in raster scan order. The block-level DCT coefficient $\lambda$ for this scheme is then:

$$\lambda = \sum_{i=2}^{N^2} c_i^2 \qquad \text{(M1)}$$

where $i = 2, \ldots, N^2$ such that $c_i \in C$, and having $N \times N$ as the size of the DCT block.

A slight variation to the scheme in method (M1), replaces the squared term in the block-level DCT coefficient function with an absolute value, which is easier to compute:

$$\lambda = \sum_{i=2}^{N^2} \left| c_i \right| \qquad \text{(M2)}$$

where $i = 2, \ldots, N^2$ such that $c_i \in C$, and having $N \times N$ as the size of the DCT block.

**Direct Accumulation with Averaging**

The algorithms introduced in this sub-section are biologically motivated by the concept of the CSF. In the HVS, receptive fields of the visual cortex display the highest sensitivity to the mid spatial frequency range. The algorithms in this section also perform high-level energy compaction, however, block coefficients generated by these methods can be considered as peaks in spatial energy corresponding to the middle of the spatial frequency spectrum. Biasing towards the middle of the frequency spectrum and energy compaction is handled by a summation of coefficient magnitudes over the DCT block, and then taking the average. This focus on the middle of the frequency spectrum is somewhat analogous to way spatial frequencies are biologically processed. we let $C$ be the set of DCT coefficients of a $N \times N$ block arranged in raster scan order. The block-level DCT coefficient $\lambda$ for this scheme is then:

$$\lambda = \frac{\sum_{i=2}^{N^2} c_i^2}{N^2 - 1} \quad \text{(M3)}$$

where $i = 2, \ldots, N^2$ such that $c_i \in C$, and having $N \times N$ as the size of the DCT block. As in the direct accumulation scheme, a variation to method (M3) can be introduced to replace the squared term in the block-level DCT coefficient function with an absolute value.

$$\lambda = \frac{\sum_{i=2}^{N^2} |c_i|}{N^2 - 1} \quad \text{(M4)}$$

where $i = 2, \ldots, N^2$ such that $c_i \in C$, and having $N \times N$ as the size of the DCT block.

**Average Absolute Deviation**

Another biologically motivated method for energy compaction and biasing towards the middle of the frequency spectrum is based on DCT block variance-based features. The average absolute deviation of the luminance coefficients from the mean coefficient value in an DCT block is indicative of the overall energy in that block, which we claim to be useful for use in face recognition. As a byproduct of the average absolute deviation computation, the block coefficient becomes partially invariant to variations (outliers) in energy of the block. The degree of invariance can be modified by adjusting the size of the block.

$$\lambda = \frac{1}{N^2 - 1} \left( \sum_{i=2}^{N^2} |c_i - \mu| \right) \quad \text{(M5)}$$

where $\mu$ is the mean of the DCT coefficients in the N x N block, computed by

$$\mu = \frac{\sum_{i=2}^{N^2} c_i}{N^2 - 1}$$

and $i = 2, \ldots, N^2$ such that $c_i \in C$, where $N \times N$ is the size of the DCT block.

**Hierarchically Superior Backpropagation Network**

In a general view of the proposed NoN system, Level 2 consists of a hierarchically superior network that classifies the information from each of the local networks. The specific network architecture chosen was the Backpropagation (BP) neural network model. The BP algorithm has been shown to work well in a variety neural network based applications. As this is the case, BP learning is employed as the classification algorithm used by the Level 2 hierarchically superior network in the NoN model (See Figure 5).

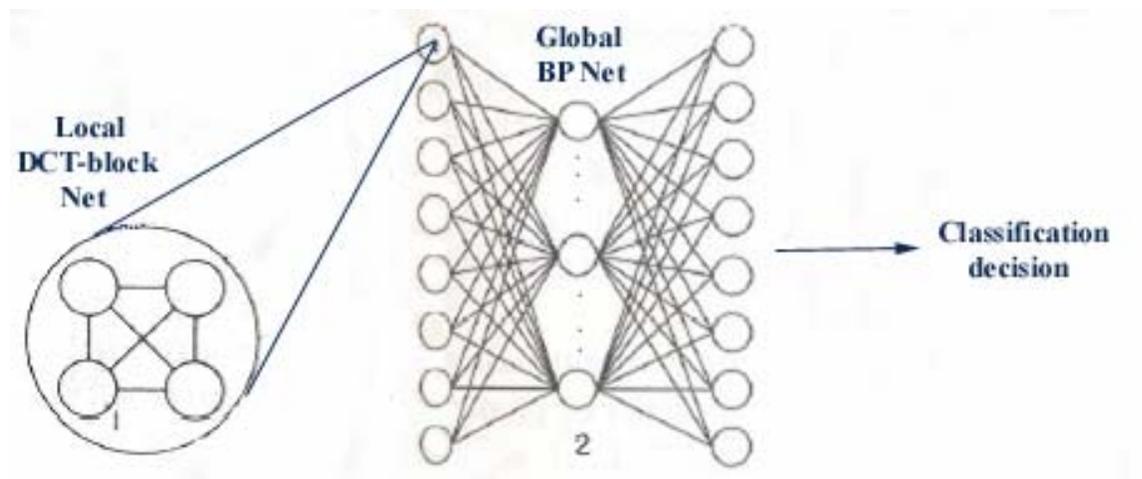

Figure 5. NoN model view of the face recognition system.

Backpropagation theory suggests that we can speed up the training procedure if all input and output vectors are in the same range. In the proposed approach, inputs and outputs are scaled to lie in the [0, 1] range. The number of outputs is the number of face subjects.

# Experimental Results

Given in this section are the results obtained from 60 different system configurations. Specifically, we provide tabular results of recognition over all Level 1 energy compaction algorithms (See Tables 1, 2, 3, 4, and 5). The average error in recognition is computed over five runs of a given network configuration. The light gray denotes the minimum average error and the dark gray represents minimum classification error for a given method. If the minimum classification error and minimum average classification error were obtained due to a single network configuration, only one row in the table is highlighted.

Table 6 is a comparison of face recognition error rates achieved when performing testing on the AT&T Cambridge Laboratories face database. The AT&T Cambridge Laboratories face database (formerly the ORL face database), was built at the Olivetti Research Laboratory in Cambridge, is available free of charge from http://www.uk.research.att.com/facedatabase.html. The database consists of 400 different images, 10 for each of 40 distinct subjects. There are 4 female and 36 male subjects. For some subjects, the images were taken at different times, varying the lighting, facial expression (open/closed eyes, smiling/not smiling) and facial details (glasses/no glasses). All the images were taken against a dark homogeneous background with the subjects in an upright, frontal position with tolerance for limited side movement and limited tilt up to about 20 degrees. There is some variation in scale of up to about 10%. The size of each image is 92x112 pixels, with 256 grey levels per pixel. Sample thumbnails of images in the AT&T Cambridge Laboratories face database can be seen in Figure 6.

In all works cited, when performing testing on the AT&T Cambridge Laboratories face database, the first five images of an individual were chosen for training, and the other five for testing (i.e., a total of 200 testing images). The error percentages in Table 6 represent the best mean reported performance.

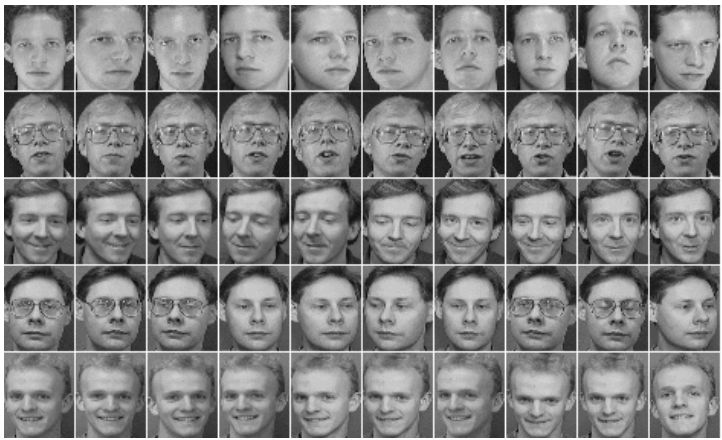

Figure 6. A sample of images in the AT&T Cambridge Laboratories face database

In the conducted experiments, the weights of the Level 2 BP network are initialized to random values in [-1.0, 1.0]. The maximum number of epochs is 300. The BP network consists of one hidden layer, with a variable number of neurons. For the AT&T Cambridge Laboratories face database, the number of outputs of the neural network is always 40. The first 5 images for each subject are the training images in total and the remaining 5 images are used for testing. There is no overlap exists between the training and test images. In each of the experiments, 5 random runs are carried out with randomly initialized weights for the BP network (for tabular results across all methods see Tables 1, 2, 3, 4 and 5).

The performance using the NoN method is very good. The mean error rate obtained was 2.5 percent and the best performance using the method was the error rate of 1.5 percent. This compares with error rates of 2.5 to 14 percent using various techniques identified in Table 6.

Table 1. Recognition rates based on method M1.

| Method | NxN | No. of Coefficients | No. of Hidden | Avg. Error (%) | Min Error (%) |
|---|---|---|---|---|---|
| M1 | 8x8 | 161 | 15 | 14.2 | 11.0 |
| | 8x8 | 161 | 25 | 11.0 | 8.5 |
| | **8x8** | **161** | **45** | **10.1** | **7.5** |
| | **8x8** | **161** | **60** | **9.0** | **8.0** |
| | 16x16 | 42 | 15 | 17.5 | 14.5 |
| | 16x16 | 42 | 25 | 12.7 | 10.5 |
| | 16x16 | 42 | 45 | 11.1 | 9.0 |
| | 16x16 | 42 | 60 | 10.9 | 10.5 |
| | 32x32 | 12 | 15 | 14.5 | 12.5 |
| | 32x32 | 12 | 25 | 16.1 | 14.5 |
| | 32x32 | 12 | 45 | 16.4 | 14.0 |
| | 32x32 | 12 | 60 | 17.4 | 17.0 |

Table 2. Recognition rates based on method M2.

| Method | NxN | No. of Coefficients | No. of Hidden | Avg. Error (%) | Min Error (%) |
|---|---|---|---|---|---|
| M2 | 8x8 | 161 | 15 | 11.5 | 10.5 |
| | 8x8 | 161 | 25 | 8.0 | 6.0 |
| | **8x8** | **161** | **45** | **4.0** | **4.5** |
| | **8x8** | **161** | **60** | **4.4** | **3.5** |
| | 16x16 | 42 | 15 | 12.0 | 9.0 |
| | 16x16 | 42 | 25 | 8.9 | 7.0 |
| | 16x16 | 42 | 45 | 7.9 | 7.5 |
| | 16x16 | 42 | 60 | 6.9 | 4.5 |
| | 32x32 | 12 | 15 | 8.1 | 6.5 |
| | 32x32 | 12 | 25 | 10.1 | 7.5 |
| | 32x32 | 12 | 45 | 12.3 | 10.0 |
| | 32x32 | 12 | 60 | 12.8 | 10.5 |

Table 3. Recognition rates based on method M3.

| Method | NxN | No. of Coefficients | No. of Hidden | Avg. Error (%) | Min Error (%) |
|---|---|---|---|---|---|
| M3 | 8x8 | 161 | 15 | 14.3 | 13.5 |
| | 8x8 | 161 | 25 | 11.7 | 8.0 |
| | 8x8 | 161 | 45 | 9.9 | 9.5 |

| | 8x8 | 161 | 60 | 8.7 | 7.0 |
| --- | --- | --- | --- | --- | --- |
| | 16x16 | 42 | 15 | 14.1 | 12.5 |
| | 16x16 | 42 | 25 | 12.1 | 10.5 |
| | 16x16 | 42 | 45 | 13.2 | 11.5 |
| | 16x16 | 42 | 60 | 10.9 | 9.5 |
| | 32x32 | 12 | 15 | 14.3 | 13.0 |
| | 32x32 | 12 | 25 | 16.0 | 14.0 |
| | 32x32 | 12 | 45 | 16.2 | 14.0 |
| | 32x32 | 12 | 60 | 17.9 | 17.5 |

Table 4. Recognition rates based on method M4.

| Method | NxN | No. of Coefficients | No. of Hidden | Avg. Error (%) | Min Error (%) |
| --- | --- | --- | --- | --- | --- |
| M4 | 8x8 | 161 | 15 | 10.5 | 7.0 |
| | 8x8 | 161 | 25 | 9.0 | 7.0 |
| | 8x8 | 161 | 45 | 4.2 | 4.0 |
| | **8x8** | **161** | **60** | **3.9** | **3.0** |
| | 16x16 | 42 | 15 | 10.5 | 8.0 |
| | 16x16 | 42 | 25 | 9.0 | 8.0 |
| | 16x16 | 42 | 45 | 6.6 | 4.0 |
| | 16x16 | 42 | 60 | 7.1 | 4.5 |
| | 32x32 | 12 | 15 | 9.6 | 8.5 |
| | 32x32 | 12 | 25 | 9.8 | 7.0 |
| | 32x32 | 12 | 45 | 12.2 | 11.0 |
| | 32x32 | 12 | 60 | 14.4 | 11.5 |

Table 5. Recognition rates obtained when using method M5. Here † Denotes minimum error over all methods and network configurations.

| Method | NxN | No. of Coefficients | No. of Hidden | Avg. Error (%) | Min Error (%) |
| --- | --- | --- | --- | --- | --- |
| M5 | 8x8 | 161 | 15 | 12.8 | 11.5 |
| | 8x8 | 161 | 25 | 6.0 | 2.5 |
| | 8x8 | 161 | 45 | 4.7 | 3.0 |
| | **8x8** | **161** | **60** | **2.6** | **1.5 †** |
| | 16x16 | 42 | 15 | 12.8 | 8.5 |
| | 16x16 | 42 | 25 | 9.2 | 7.0 |
| | 16x16 | 42 | 45 | 4.8 | 4.5 |
| | 16x16 | 42 | 60 | 7.9 | 6.0 |
| | 32x32 | 12 | 15 | 9.0 | 6.5 |
| | 32x32 | 12 | 25 | 9.9 | 8.5 |
| | 32x32 | 12 | 45 | 12.9 | 10.5 |
| | 32x32 | 12 | 60 | 12.1 | 11.0 |

Table 6. Comparison of average error rates in current literature when using the AT&T Cambridge Laboratories face database.

| System | Error (%) |
| --- | --- |
| Gaussian Weighting [5] | 14.0 |
| Hidden Markov Models [10] | 14.0 |
| Eigenface [13] | 10.0 |
| Linear Discriminant Analysis [17] | 9.2 |
| Low frequency DCT with subimages [11] | 7.35 |
| Low frequency DCT w/o subimages [12] | 4.85 |
| Pseudo-2D Hidden Markov Models [14] | 4.0 |
| Convolutional neural network [8] | 3.8 |
| Linear support vector machines [4] | 3.0 |
| Block-level DCT Coefficient Features (NoN) | 2.5 |
| Kernel PCA [7] | 2.5 |
| One Spike Neural Network [2] | 2.5 |
| Uncorrelated Discriminant Transform [18] | 2.5 |

# Concluding Remarks

The proposed NoN model for face recognition is motivated by biological information processing, namely cortical processing and spatial vision in the HVS. In comparison to other biologically inspired models, the NoN model performs quite well, but its limitations [19-21] should not be overlooked. The proposed approach has been shown to be a method of autonomic face recognition worthy of future study.

While the proposed method exhibits promise as a model for biological processes in human face recognition, this approach suffers from certain limitations. One drawback is its susceptibility to blocking artifact degradation, due to the artificial discontinuities that appear between the boundaries of the blocks, and these artifacts remain typically the greatest form of image degradation in block transform coding systems. Another drawback is the need to zero-pad images to make them divisible by the subimage blocks, and the introduction of this redundant data reduces classification accuracy. Yet another drawback of the approach is the computational cost incurred in computing the block-level coefficient features in Level 1 of the NoN model. While there is a computational benefit to applying the $N \times N$ 2-D DCT to an image, this benefit is reduced by performing energy compaction algorithms, based on summations, over the DCT block.

While the proposed NoN system performed well, modifications to the system may improve recognition. One possible area of investigation is representing face images as nonuniformly sampled data points. Another potential area of research is alternative algorithms at both levels 1 and 2 of the NoN model. Although the set of Level 1 algorithms introduced in this dissertation performed well, and the computations had a biological basis, recognition rates may improve using alternate algorithms. Currently, the superior classifier used is a backpropagation (BP) neural network, but the popular BP classification has its drawbacks, therefore new classes of instantaneously trained neural network algorithms may be investigated [6].

Another area of future work involves investigating advanced NoN models. The proposed 2-level NoN model has full connectivity as the higher level uses all the DCT-based coefficients simultaneously. It would be worthwhile to investigate systems where only spatially adjacent blocks are connected at the higher level. Furthermore, three or higher-level systems may also be investigated. In a three-level system, the block size at the second level could be larger than the size at the first level. It would be useful to study the effect on performance of a system where the block sizes are adaptively adjusted in the learning phase.